\documentclass[11pt]{article}

\usepackage[final]{acl}

\usepackage{times}
\usepackage{latexsym}

\usepackage[T1]{fontenc}
\usepackage[utf8]{inputenc}

\usepackage{microtype}

\usepackage{inconsolata}

\usepackage{graphicx}
\usepackage{booktabs}
\usepackage{amsmath}
\usepackage{amssymb}
\usepackage{xcolor}
\usepackage{tcolorbox}
\usepackage{amsmath}
\usepackage{multirow}
\usepackage{pifont}
\usepackage{makecell}

\newcommand{\villainemoji}{%
  \raisebox{-0.2em}{\includegraphics[height=1.2em]{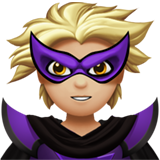}}%
}

\usepackage{kotex}

\title{\villainemoji~VILLAIN at AVerImaTeC:\\Verifying Image–Text Claims via Multi-Agent Collaboration}

\author{
Jaeyoon Jung$^{\clubsuit\diamondsuit}$~~~~~~Yejun Yoon$^{\heartsuit}$~~~~~~Kunwoo Park$^{\clubsuit\heartsuit}$\\ 
$^{\clubsuit}$School of AI Convergence, Soongsil University\\
$^{\diamondsuit}$MAUM AI Inc.\\
$^{\heartsuit}$Department of Intelligent Semiconductors, Soongsil University\\
\texttt{\{jaeyoonskr, yejun0382\}@soongsil.ac.kr}, \texttt{kunwoo.park@ssu.ac.kr}
}

\begin{document}
\maketitle
\begin{abstract}
This paper describes VILLAIN, a multimodal fact-checking system that verifies image--text claims through prompt-based multi-agent collaboration. For the AVerImaTeC shared task, VILLAIN employs vision--language model agents across multiple stages of fact-checking. Textual and visual evidence is retrieved from the knowledge store enriched through additional web collection. To identify key information and address inconsistencies among evidence items, modality-specific and cross-modal agents generate analysis reports. In the subsequent stage, question--answer pairs are produced based on these reports. Finally, the Verdict Prediction agent produces the verification outcome based on the image-text claim and the generated question--answer pairs. Our system ranked first on the leaderboard across all evaluation metrics. The source code is publicly available at \url{https://github.com/ssu-humane/VILLAIN}.

\end{abstract}

\section{Introduction}

Motivated by the spread of misleading multimodal content and the need for automated verification, the AVerImaTeC shared task at the 9th FEVER Workshop focuses on verifying real-world image–text claims using external evidence~\cite{cao2025averimatec}, reflecting web-based fact-checking scenarios in the wild. Given a claim and its metadata, systems must retrieve relevant evidence from the web or from a provided document and image collection and assess the claim's validity. Claims are labeled as Supported, Refuted, Not Enough Evidence, or Conflicting Evidence/Cherry-picking, and a prediction is considered correct only if both the label and the accompanying evidence are appropriate.

This paper describes \villainemoji~\textbf{VILLAIN}, developed by Team HUMANE, a fact-checking system designed to verify image--text claims using vision--language model (VLM) agents. The system employs agents operating across multiple stages of automated fact-checking. The first two stages, Evidence Retrieval and Analysis, aim to identify evidence that facilitates veracity prediction. Using state-of-the-art embedding and rerankers, we retrieve textual and visual evidence from three knowledge stores, which are provided with the original dataset and further enriched through additional web content collection. To address noise and inconsistencies among evidence items, we employ three agents that generate analysis reports. The subsequent stage is Question--Answer Generation, in which an agent produces question--answer pairs by synthesizing the reports from the previous stage. Finally, in the Verdict Prediction stage, an agent predicts a verdict label together with a textual justification and the selected question--answer pairs. Our system achieved first place on the leaderboard, obtaining a veracity score of 0.546 on the test set and outperforming all other systems across the official evaluation metrics.

\begin{figure*}[t]
    \centering
    \includegraphics[width=.99\linewidth]{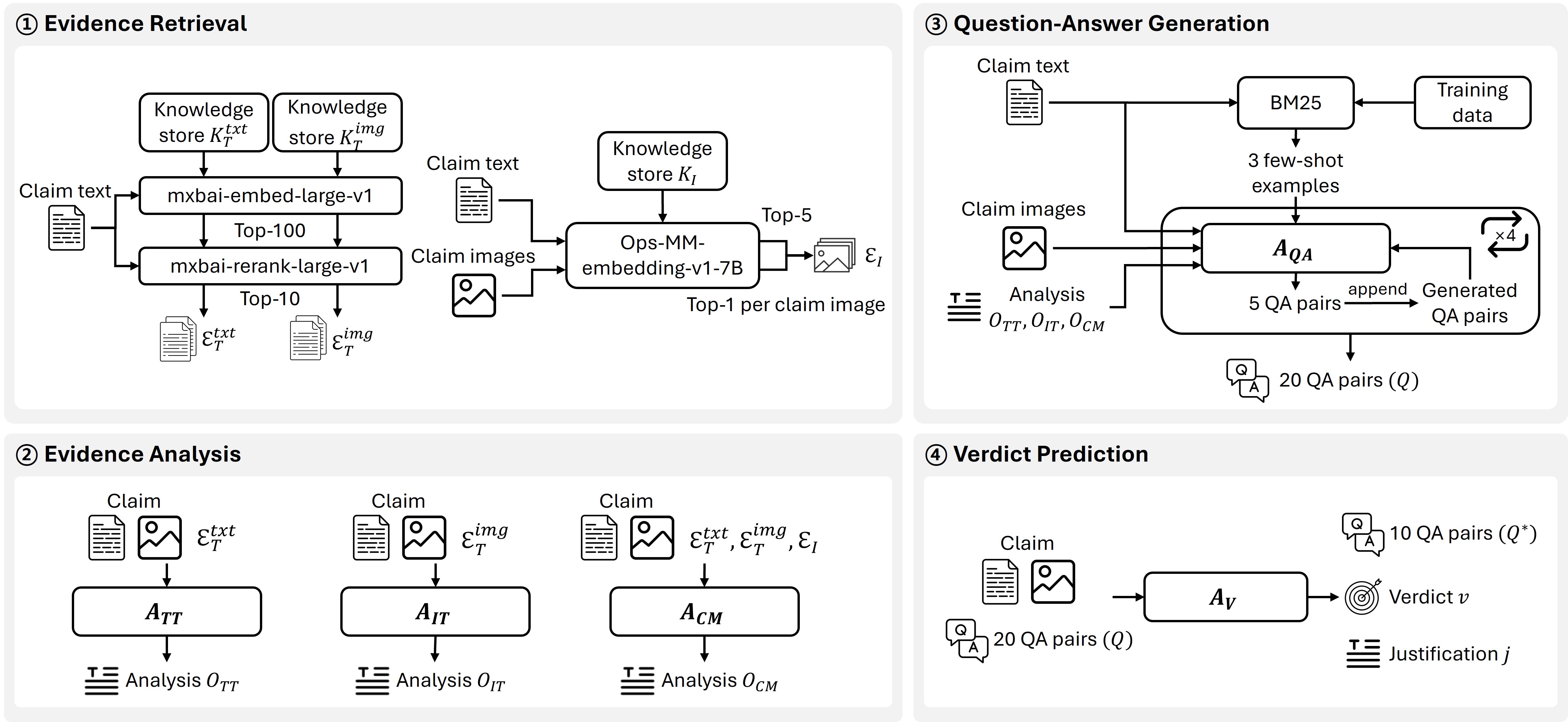}
    \caption{Overview of \villainemoji~\textbf{VILLAIN}.}
    \label{fig:pipeline}
\end{figure*}

\section{Related Works}

AVeriTeC~\cite{schlichtkrull2023averitec} is a text-only fact-checking dataset that was used in the two previous shared tasks hosted by the FEVER workshop. At the Seventh FEVER Workshop~\cite{schlichtkrull-etal-2024-automated}, \citet{rothermel2024infact} proposed an LLM-based claim verification framework consisting of six stages built on GPT-4o, demonstrating the strong performance of closed-source models. In contrast, \citet{yoon-etal-2024-hero} introduced HerO based on open LLMs and highlighted the importance of evidence retrieval for verification accuracy. The subsequent task hosted at the Eighth FEVER Workshop imposed additional computational constraints~\cite{akhtar-etal-2025-2nd}. Under these constraints, CTU AIC proposed a two-stage pipeline based on long-context retrieval-augmented generation~\cite{ullrich2025aic} to capture salient sentences and paragraphs. HerO 2~\cite{yoon-etal-2025-team} further improved upon HerO by applying model quantization and answer reformulation, emphasizing the importance of distilling answers to retain only the most relevant factual information.

VILLAIN is the first multi-agent system designed for AVerImaTeC, in which individual agents assume distinct roles across consecutive stages of the fact-checking process. Inspired by the strong systems proposed for earlier shared tasks, we adopt a prompt-based design that enables experimentation with both powerful open- and closed-source models, enhances the retrieval stage through reranking and knowledge store filling, and employs dedicated evidence analysis agents to identify and distill the most informative evidence for verification.

\section{Task Description}

The AVerImaTeC shared task focuses on developing automated multimodal fact-checking systems capable of verifying real-world image--text claims using retrieved web evidence. Given an image-text claim along with its associated metadata, a fact-checking system is required to generate a verdict. The supporting evidence may be retrieved either directly from the web or from the document and image evidence collections provided by the organizers. A response is considered correct only if both the predicted label is correct and the provided evidence is adequate. Specifically, a verdict is deemed correct only when the evidence score meets or exceeds the threshold $\lambda=0.3$. Further details can be found in \citet{cao2025averimatec}.

\section{Our System}

This section describes \villainemoji~\textbf{VILLAIN} (\textbf{V}erifying \textbf{I}mage-text c\textbf{L}aims via Mu\textbf{L}ti-\textbf{A}gent collaborat\textbf{I}o\textbf{N}), developed by Team HUMANE, a multi-agent multimodal fact-checking pipeline that verifies claims through collaborative VLM agents. As illustrated in Figure~\ref{fig:pipeline}, VILLAIN verifies a claim through a four-step sequential pipeline. Gemini-2.5-Pro~\cite{comanici2025gemini} is selected as the base model for all VLM agents.

\subsection{Evidence Retrieval}

The first stage aims to retrieve web evidence to verify a given claim. We use the set of knowledge stores provided with AVerImaTeC~\cite{cao2025averimatec}, denoted as $\mathcal{K}=\{\mathcal{K}^{txt}_{T},\mathcal{K}^{img}_{T},\mathcal{K}_{I}\}$. Each of the three stores contains a distinct type of evidence. Specifically, $\mathcal{K}^{txt}_{T}$ denotes the knowledge store consisting of web text retrieved using a textual query; $\mathcal{K}^{img}_{T}$ denotes the knowledge store comprising textual evidence obtained via reverse image search applied to the claim images; and $\mathcal{K}_{I}$ denotes the knowledge store of web images retrieved using a textual query.

From a preliminary investigation of the data, we found that only URLs were provided for a substantial portion of samples in $\mathcal{K}_{T}$, while the corresponding text fields were missing. In addition, some evidence entries contained only generic website components (e.g., navigation bars, footers, or cookie notices) or access-restricted content (e.g., login walls) and therefore did not provide meaningful information for verification.

To address these issues, we perform an web content extraction process using Playwright\footnote{\url{https://github.com/microsoft/playwright}}, a browser automation framework, to populate knowledge-store entries that were previously empty or contained only generic or access-restricted content. A headless Chromium browser is employed to mitigate access restrictions that commonly affect non-browser clients. Knowledge-store entries are crawled either when their content is missing or when the existing content fails a usefulness check based on content length and the proportion of generic patterns. Table~\ref{tab:evidence-filling} reports statistics on the number of filled evidence items per claim before and after the crawling process. Further details can be found at Appendix~\ref{sec:appendix-knowledge-store}.

\paragraph{Textual Evidence Retrieval}
We retrieve textual evidence from both $\mathcal{K}^{txt}_{T}$ and $\mathcal{K}^{img}_{T}$ using mxbai-embed-large-v1~\cite{emb2024mxbai} as the text embedding model and mxbai-rerank-large-v1~\cite{rerank2024mxbai} as the text reranking model. For each knowledge store, we first retrieve the top-100 candidate evidence based on embedding similarity and then select top-10 evidence by reranking them. Following the winning approach of the previous AVeriTeC shared task~\cite{ullrich-drchal-2025-aic}, we segment each document into chunks of 2,048 characters. After retrieval, we augment each selected chunk with its adjacent chunks (one preceding and one following) to incorporate local context, yielding an evidence item consisting of three consecutive chunks. We denote the retrieved textual evidence sets as $\mathcal{E}^{txt}_{T}$ and $\mathcal{E}^{img}_{T}$, respectively.

\paragraph{Visual Evidence Retrieval}
We retrieve visual evidence from $\mathcal{K}_{I}$ using Ops-MM-embedding-v1-7B, a multimodal embedding model. Specifically, we retrieve the top-5 visual evidence items using the claim text as the query and retrieve the top-1 visual evidence item for each claim image. We denote the retrieved visual evidence set as $\mathcal{E}_{I}$.

\subsection{Evidence Analysis}
Some of the retrieved evidence may be irrelevant to the claim or lead to inconsistencies across different evidence items. To address these issues, we employ three VLM-based analysis agents, each responsible for analyzing a different set of retrieved evidence in the previous stage. 

\begin{table}[t]
\centering
\small
\begin{tabular}{lllrrr}
\toprule
\textbf{Split} & \textbf{Store} & \textbf{Status} & \textbf{Avg} & \textbf{Min} & \textbf{Max} \\
\midrule
\multirow{4}{*}{Dev} 
  & \multirow{2}{*}{$\mathcal{K}^{txt}_{T}$} 
      & Original & 654.3 & 102 & 1,277 \\
  &   & Filled   & 837.5 & 133 & 1,683 \\
\cmidrule{2-6}
  & \multirow{2}{*}{$\mathcal{K}^{img}_{T}$} 
      & Original & 14.5 & 1 & 70 \\
  &   & Filled   & 16.3 & 1 & 91 \\
\midrule
\multirow{4}{*}{Test} 
  & \multirow{2}{*}{$\mathcal{K}^{txt}_{T}$} 
      & Original & 1,429.0 & 160 & 2,900 \\
  &   & Filled   & 1,788.8 & 196 & 3,551 \\
\cmidrule{2-6}
  & \multirow{2}{*}{$\mathcal{K}^{img}_{T}$} 
      & Original & 23.5 & 1 & 174 \\
  &   & Filled   & 27.2 & 1 & 266 \\
\bottomrule
\end{tabular}
\caption{Statistics of filled evidence per claim before and after URL content filling. Each URL with non-empty text content is counted as one evidence item.}
\label{tab:evidence-filling}
\end{table}

Specifically, the Text-Text agent ($\mathcal{A}_{TT}$) analyzes textual evidence associated with the claim text ($\mathcal{E}^{txt}_{T}$), while the Image-Text agent ($\mathcal{A}_{IT}$) examines textual evidence derived from the claim images ($\mathcal{E}^{img}_{T}$). The Cross-Modal agent ($\mathcal{A}_{CM}$) investigates relationships across both textual and visual evidence ($\mathcal{E}_{I}$). The analysis process is formalized as follows:
\begin{align}
    \mathcal{O}_{TT} &= \mathcal{A}_{TT}(c, \mathcal{E}^{txt}_{T} ; \mathcal{P}_{TT}), \\
    \mathcal{O}_{IT} &= \mathcal{A}_{IT}(c, \mathcal{E}^{img}_{T} ; \mathcal{P}_{IT}), \\
    \mathcal{O}_{CM} &= \mathcal{A}_{CM}(c, \mathcal{E}^{txt}_{T}, \mathcal{E}^{img}_{T}, \mathcal{E}_{I} ; \mathcal{P}_{CM}),
\end{align}
where $\mathcal{O}$ denotes the analysis output generated by each agent and $\mathcal{P}_k$ ($k\in \{TT, IT, CM\}$) denotes the prompt used (shown in Figure~\ref{fig:agent1-prompt},~\ref{fig:agent2-prompt}, and \ref{fig:agent3-prompt}).

\subsection{Question--Answer Generation}

The Question--Answer (QA) generation agent $\mathcal{A}_{QA}$ synthesizes the preliminary analysis outputs $\mathcal{O}_{TT}$, $\mathcal{O}_{IT}$, and $\mathcal{O}_{CM}$ from the previous stage into QA pairs using a VLM. We generated five QA pairs iteratively by conditioning on the previously generated QA pairs. This design facilitates the generation of coherent follow-up question-answer pairs, while reducing redundancy by preserving context from the QA generation history.

Formally, the QA generation process at iteration $i$ is defined as:
\begin{equation}
\mathcal{Q}^{(i)} = \mathcal{A}_{QA}\big(c,\,\mathcal{O}_{TT},\,\mathcal{O}_{IT},\,\mathcal{O}_{CM},\,\mathcal{Q}^{(<i)}; \mathcal{P}_{QA}\big),
\end{equation}
where $\mathcal{P}_{QA}$ denotes the task-specific prompt for QA generation (shown in Figure~\ref{fig:agent4-prompt}), $\mathcal{Q}^{(i)}$ represents the QA pairs generated at iteration $i$, and $\mathcal{Q}^{(<i)}$ denotes the set of all QA pairs generated prior to iteration $i$. We employ in-context learning with retrieved three few-shot examples from the training data using BM25, and these examples are incorporated into the QA generation prompt $\mathcal{P}_{QA}$. In our implementation, QA generation is conducted over four iterations, with five QA pairs generated at each iteration. Newly generated QA pairs are appended to the existing set and fed back into the model as part of the input for the next iteration, resulting in a total of 20 QA pairs. We denote the whole set of generated QA pairs as $\mathcal{Q} = \cup_{i=1}^{4} \mathcal{Q}^{(i)}$.

\subsection{Verdict Prediction}

In the final stage, the Verdict Prediction agent $\mathcal{A}_{V}$ verifies the claim. Given $\mathcal{Q}$, the set of QA pairs generated in the previous stage, $\mathcal{A}_{V}$ is instructed to predict a verdict label along with a textual justification while selecting the top-10 most relevant QA pairs.

The verdict prediction process is formally defined as:
\begin{equation}
(v, j, \mathcal{Q}^{*}) = \mathcal{A}_{V}\big(c,\,\mathcal{Q}; \mathcal{P}_{V}\big),
\end{equation}
where $v$ denotes the predicted veracity label, $j$ denotes the corresponding textual justification, $\mathcal{P}_{V}$ denotes the prompt used (shown in Figure~\ref{fig:agent5-prompt}), and $\mathcal{Q}^{*}$ denotes the subset of $\mathcal{Q}$ selected by $\mathcal{A}_{V}$. 

\section{Evaluation Experiments}

\subsection{Experimental Setups}
We evaluate model performance on the development set and conduct experiments with different vision–language model backbones, including the Gemini-2.5~\cite{comanici2025gemini} and Qwen3-VL~\cite{Qwen3-VL} model series, both of which incorporate inherent reasoning capabilities.

We evaluate our pipeline using Gemma-3-27B~\cite{team2025gemma}, the model used for evaluation in the official AVerImaTeC shared task repository\footnote{\url{https://github.com/abril4416/AVerImaTec_Shared_Task}}. Each evaluation is repeated five times to account for variability across runs, and we report the mean scores along with the standard deviation. As specified by the AVerImaTeC evaluation protocol, we report four metrics:
\begin{itemize}
    \item \textsf{Q-Eval}: A reference-based recall metric for comparing the generated questions with the gold questions.
    \item \textsf{Evid-Eval}: A reference-based recall metric for comparing the system evidence with the gold evidence, computed via textual matching followed by image matching. 
    \item \textsf{Veracity}: The conditional accuracy of the predicted verdict label.
    \item \textsf{Justification}: A reference-based semantic similarity metric computed using Ev2R~\cite{akhtar2024ev2r}, which compares the generated justification with the human-annotated justification.
\end{itemize}

Both Veracity and Justification are reported conditionally, considering only instances for which the associated evidence score exceeds a predefined threshold $\lambda$. Instances falling below this threshold are treated as incorrect. We report Veracity and Justification score under both conditioned settings ($\lambda = 0.3$, the final evaluation setting of the shared task) and the unconditioned ($\lambda = 0$) settings.

We conducted our experiments on two machines: one equipped with eight H100 GPUs and 2~TB of RAM, and another with four NVIDIA A6000 GPUs and 256~GB of RAM. All experiments were performed in a computing environment configured with Python 3.11.13, PyTorch 2.7.0, Transformers 4.57.3, and SentenceTransformers 5.1.2.

\subsection{Experimental Results}

\begin{table}[t]
\centering
\resizebox{0.99\linewidth}{!}{
\begin{tabular}{lcccc}
\toprule
System &  Q-Eval & Evid-Eval & Veracity & Justification \\\midrule
HUMANE (\villainemoji) & \textbf{0.923} & \textbf{0.583} & \textbf{0.645} & \textbf{0.543} \\
ADA-AGGR & 0.354 & 0.386 & 0.454 & 0.373 \\
AIC CTU & 0.822 & 0.347 & 0.375 & 0.304 \\
\multicolumn{5}{c}{\emph{(Additional systems omitted for brevity)}} \\
Baseline & 0.488 & 0.134 & 0.066 & 0.058 \\
\bottomrule
\end{tabular}}
\caption{Leaderboard results on the development set.}
\label{tab:main_results}
\end{table}

\begin{table}[t]
\centering
\resizebox{0.99\linewidth}{!}{
\begin{tabular}{lcccc}
\toprule
System &  Q-Eval & Evid-Eval & Veracity & Justification \\\midrule
HUMANE (\villainemoji) & \textbf{0.890} & \textbf{0.536} & \textbf{0.546} & \textbf{0.556} \\
ADA-AGGR & 0.370 & 0.463 & 0.537 & 0.433 \\
AIC CTU & 0.807 & 0.325 & 0.347 & 0.304 \\
Baseline & 0.555 & 0.171 & 0.114 & 0.132 \\
\bottomrule
\end{tabular}}
\caption{Leaderboard results on the test set.}
\label{tab:test_results}
\end{table}

\paragraph{Development Set Results}

Table~\ref{tab:main_results} presents the performance of VILLAIN on the development set, in comparison with the baseline and other competitive systems evaluated on the official Hugging Face leaderboard\footnote{\url{https://huggingface.co/spaces/FEVER-IT/AVerImaTeC}}. VILLAIN achieves the highest veracity accuracy of 0.645, followed by ADA-AGGR (0.454) and AIC CTU (0.375). The performance gap between VILLAIN and the baseline is substantial. Moreover, VILLAIN attains the best scores across Q-EVAL, Evid-Eval, and Justification, indicating its effectiveness throughout multiple stages of the fact-checking pipeline. 

\paragraph{Test Set Results}
Table~\ref{tab:test_results} reports the test set performance of VILLAIN in comparison with the baseline and other top-ranked submissions. Our system achieves the highest veracity score (0.546), outperforming ADA-AGGR (0.537) and AIC CTU (0.347). Moreover, VILLAIN obtains the best scores across all evaluation metrics reflecting each stage of fact-checking.

\paragraph{Ablation Study: Evidence Analysis}

We conduct an ablation study to assess the contribution of the evidence analysis agents by removing them and directly passing the retrieved evidence to the QA generation agent. Table~\ref{tab:evidence-agent-ablation} compares model performance with and without evidence analysis agents. The results show that incorporating evidence analysis agents consistently improves the Evid-Eval scores across all evaluated models, with gains of 0.040 for Gemini-2.5-Flash, 0.067 for Qwen3-VL-32B, and 0.043 for Qwen3-VL-8B. These improvements suggest that the evidence analysis agents help identify and surface more relevant evidence, which in turn contributes to improved verdict prediction accuracy. 

\begin{table}[t]
\centering
\resizebox{0.99\linewidth}{!}{
\begin{tabular}{lcccc}
\toprule
Model &
\makecell{Evidence \\ Analysis} &
Q-Eval &
Evid-Eval &
Veracity \\
\midrule
\multirow{2}{*}{Gemini-2.5-Flash}
    & \ding{51}  & 0.841$_{\pm.006}$ & 0.511$_{\pm.016}$ & 0.554$_{\pm.025}$ \\
    & \ding{55}  & 0.858$_{\pm.014}$ & 0.471$_{\pm.018}$ & 0.516$_{\pm.036}$ \\
\midrule
\multirow{2}{*}{Qwen3-VL-32B}
    & \ding{51}  & 0.803$_{\pm.009}$ & 0.489$_{\pm.017}$ & 0.532$_{\pm.022}$ \\
    & \ding{55}  & 0.848$_{\pm.011}$ & 0.422$_{\pm.017}$ & 0.483$_{\pm.023}$ \\
\midrule
\multirow{2}{*}{Qwen3-VL-8B}
    & \ding{51}  & 0.737$_{\pm.008}$ & 0.396$_{\pm.008}$ & 0.421$_{\pm.015}$ \\
    & \ding{55}  & 0.762$_{\pm.012}$ & 0.353$_{\pm.021}$ & 0.386$_{\pm.025}$ \\
\bottomrule
\end{tabular}
}
\caption{Ablation study on the impact of evidence analysis agents. \ding{55} denotes disabling the evidence analysis agents and directly passing retrieved evidence to the QA generation agent, while \ding{51} denotes the full system with evidence analysis agents enabled.}
\label{tab:evidence-agent-ablation}
\end{table}

\paragraph{Ablation Study: Knowledge Store Filling}
We evaluate the impact of our filled knowledge store on task performance. Table~\ref{tab:knowledge-store-ablation} compares model performance when using the original knowledge store versus the filled knowledge store that has been.

Filling the knowledge store consistently improves Evid-Eval scores across all models, with gains of 0.03 for Gemini-2.5-Flash, 0.038 for Qwen3-VL-32B, and 0.036 for Qwen3-VL-8B. The improvements in question generation quality may be attributed to richer evidence, which can help models formulate more relevant diagnostic questions. Notably, Qwen3-VL-8B also shows a 0.033 improvement in verdict accuracy, indicating that additional evidence can help smaller models make more accurate predictions.

\begin{table}[t]
\centering
\resizebox{0.99\linewidth}{!}{
\begin{tabular}{lcccc}
\toprule
Model &
Fill &
Q-Eval &
Evid-Eval &
Veracity \\
\midrule
\multirow{2}{*}{Gemini-2.5-Flash}
    & \ding{51} & 0.841$_{\pm.006}$ & 0.511$_{\pm.016}$ & 0.554$_{\pm.025}$ \\
    & \ding{55} & 0.817$_{\pm.008}$ & 0.481$_{\pm.016}$ & 0.538$_{\pm.018}$ \\
\midrule
\multirow{2}{*}{Qwen3-VL-32B}
    & \ding{51} & 0.803$_{\pm.009}$ & 0.489$_{\pm.017}$ & 0.532$_{\pm.022}$ \\
    & \ding{55} & 0.782$_{\pm.005}$ & 0.451$_{\pm.010}$ & 0.505$_{\pm.015}$ \\
\midrule
\multirow{2}{*}{Qwen3-VL-8B}
    & \ding{51} & 0.737$_{\pm.008}$ & 0.396$_{\pm.008}$ & 0.421$_{\pm.015}$ \\
    & \ding{55} & 0.740$_{\pm.015}$ & 0.339$_{\pm.011}$ & 0.388$_{\pm.017}$ \\
\bottomrule
\end{tabular}
}
\caption{Effect of knowledge store filling on model performance. \ding{51} denotes a filled knowledge store, and \ding{55} denotes the original knowledge store.}
\label{tab:knowledge-store-ablation}
\end{table}

\paragraph{Ablation Study: Backbone VLM Comparison}
Table~\ref{tab:main-results} reports the experimental results on the development set obtained by varying the backbone VLMs used across all agent components, including evidence analysis, question generation, and verdict prediction. Specifically, we compare models from the Gemini-2.5 series (Pro and Flash) with those from the Qwen3-VL (32B, 8B, 4B, 2B) series. 

Results demonstrate a clear performance trend in which Gemini-2.5-Pro outperforms all other models across most evaluation metrics. A single exception is observed for unconditioned verdict accuracy ($\lambda=0$), where Qwen3-VL-32B performs comparably to Gemini-2.5-Pro and achieves a score that is 0.026 higher than Gemini-2.5-Flash. However, Qwen3-VL-32B yields lower scores for the other metrics, even when compared to Gemini-2.5-Flash, a more lightweight variant of Gemini-2.5-Pro.

\begin{table*}[t]
\centering
\resizebox{0.99\textwidth}{!}{
\begin{tabular}{lccccccc}
\toprule
\multirow{2}{*}{\textbf{Model}} &
\multirow{2}{*}{\textbf{Size (B)}} &
\multirow{2}{*}{\textbf{Q-Eval}} &
\multirow{2}{*}{\textbf{Evid-Eval}} &
\multicolumn{2}{c}{\textbf{Veracity}} &
\multicolumn{2}{c}{\textbf{Justification}} \\
& & & & \textbf{$\lambda = 0$} & \textbf{$\lambda = 0.3$} &
\textbf{$\lambda = 0$} & \textbf{$\lambda = 0.3$} \\
\midrule
Gemini-2.5-Pro & - & \textbf{0.913}$_{\pm.009}$ & \textbf{0.588}$_{\pm.014}$ & \textbf{0.908}$_{\pm.000}$ & \textbf{0.634}$_{\pm.023}$ & \textbf{0.757}$_{\pm.011}$ & \textbf{0.566}$_{\pm.014}$ \\
Gemini-2.5-Flash & - & 0.841$_{\pm.006}$ & 0.511$_{\pm.016}$ & 0.882$_{\pm.000}$ & 0.554$_{\pm.025}$ & 0.677$_{\pm.006}$ & 0.477$_{\pm.020}$ \\
\midrule
\multirow{4}{*}{Qwen3-VL} & 32 & 0.803$_{\pm.009}$ & 0.489$_{\pm.017}$ & \textbf{0.908}$_{\pm.000}$ & 0.532$_{\pm.022}$ & 0.650$_{\pm.010}$ & 0.441$_{\pm.020}$ \\
& 8 & 0.737$_{\pm.008}$ & 0.396$_{\pm.008}$ & 0.842$_{\pm.000}$ & 0.421$_{\pm.015}$ & 0.550$_{\pm.018}$ & 0.361$_{\pm.013}$ \\
& 4 & 0.713$_{\pm.003}$ & 0.382$_{\pm.018}$ & 0.862$_{\pm.000}$ & 0.420$_{\pm.032}$ & 0.540$_{\pm.007}$ & 0.321$_{\pm.020}$ \\
& 2 & 0.562$_{\pm.018}$ & 0.239$_{\pm.009}$ & 0.763$_{\pm.000}$ & 0.272$_{\pm.014}$ & 0.333$_{\pm.017}$ & 0.173$_{\pm.010}$ \\
\bottomrule
\end{tabular}}
\caption{Performance comparison between Qwen3-VL models of different scales and Gemini-2.5 models.}
\label{tab:main-results}
\end{table*}

\section{Conclusion}
We present \villainemoji~\textbf{VILLAIN}, the winning system in the AVerImaTeC shared task. It is a multimodal fact-checking pipeline that verifies image--text claims through prompt-based multi-agent collaboration. Using embedding-based retrieval and reranking, VILLAIN retrieves textual and visual evidence from the AVerImaTeC knowledge stores, which are further enriched with URL content filling. VLM agents perform modality-specific and cross-modal analysis, generate QA pairs, and produce a final verdict with justification and selected QA pair. Experimental results show that VILLAIN consistently outperforms competing systems across all evaluation metrics. Overall, these results highlight the effectiveness of multi-agent collaboration and iterative reasoning adopted in our system. By releasing our source code with support for multiple VLM backbones, we hope that VILLAIN will serve as a strong foundation for future research on reliable and transparent multimodal fact-checking systems.

\section*{Limitations}
\paragraph{Generalizability} While VILLAIN achieves state-of-the-art performance on AVerImaTeC, it has not been evaluated on other datasets. Therefore, the effectiveness of the system should be interpreted within the specific scope of the task and dataset. Future work could extend the dedicated approach into a more generalizable multi-agent framework for automated fact-checking.

\paragraph{Computational Costs} For each claim, VILLAIN requires eight calls to a reasoning-capable VLM, which substantially increases the number of input and output tokens and, consequently, inference costs. Following recent advancements in efficient deployment~\cite{feng2025efficient} and the potential of small models for multi-agent systems~\cite{belcak2025small}, future research could investigate a lightweight variant of the system.

\section*{Acknowledgments}

This research was supported by the IITP (Institute of Information \& Communications Technology Planning \& Evaluation), funded by the Korea government (MSIT) (IITP-2026-RS-2022-00156360, IITP-2026-RS-2024-00430997, IITP-2026-RS-2020-II201602). Jaeyoon Jung and Yejun Yoon contributed equally to this work as co-first authors. Kunwoo Park is the corresponding author.

\bibliography{custom}

\newpage
\appendix

\section{Model Specifications}

We evaluate models from the Qwen3-VL series (thinking variants) as well as the Gemini-2.5 series.

\begin{itemize}
    \item \textbf{Qwen3-VL-2B}: \url{https://huggingface.co/Qwen/Qwen3-VL-2B-Thinking}
    \item \textbf{Qwen3-VL-4B}: \url{https://huggingface.co/Qwen/Qwen3-VL-4B-Thinking}
    \item \textbf{Qwen3-VL-8B}: \url{https://huggingface.co/Qwen/Qwen3-VL-8B-Thinking}
    \item \textbf{Qwen3-VL-32B}: \url{https://huggingface.co/Qwen/Qwen3-VL-32B-Thinking}
    \item \textbf{Gemini-2.5-Flash}: \url{https://ai.google.dev/gemini-api/docs/models?hl=en#gemini-2.5-flash}
    \item \textbf{Gemini-2.5-Pro}: \url{https://ai.google.dev/gemini-api/docs/models?hl=en\#gemini-2.5-pro}
\end{itemize}

For the embedding and reranking components, we use the following models:

\begin{itemize}
    \item \textbf{mxbai-embed-large-v1}: \url{https://huggingface.co/mixedbread-ai/mxbai-embed-large-v1}
    \item \textbf{mxbai-rerank-large-v1}: \url{https://huggingface.co/mixedbread-ai/mxbai-rerank-large-v1}
    \item \textbf{Ops-MM-embedding-v1-7B}: \url{https://huggingface.co/OpenSearch-AI/Ops-MM-embedding-v1-7B}
\end{itemize}

For the Gemma 3 27B model used in the evaluation, we employ:

\begin{itemize}
    \item \textbf{Gemma-3-27b}: \url{https://huggingface.co/google/gemma-3-27b-it}
\end{itemize}

\section{Knowledge Store Construction}
\label{sec:appendix-knowledge-store}

The AVerImaTeC shared task provides a knowledge store containing pre-retrieved evidence for each claim. However, we observed that many evidence entries had empty \texttt{url2text} fields, limiting the textual information available for fact-checking. To address this, we developed an automated URL content extraction pipeline to fill these missing fields.

\subsection{URL Content Extraction Methodology}

We implemented a robust web scraping pipeline that handles diverse web content types and JavaScript-heavy websites. The pipeline employs a two-tier approach:

\paragraph{Standard HTTP Requests} 
For static websites, we use the requests\footnote{\url{https://pypi.org/project/requests/}} library with custom headers to fetch HTML content. This approach is efficient for news sites, academic publications, and most standard web pages.

\paragraph{JavaScript Rendering} 
For dynamic websites that require JavaScript execution (e.g., social media platforms and archive sites), we employ Playwright with Chromium to fully render pages before extracting content. Requests to domains such as Facebook, Instagram, Twitter/X, TikTok, LinkedIn, Quora, YouTube, and web archive services are automatically routed to Playwright, as these sites typically fail to return complete content when accessed via standard HTTP requests.

We filter out login walls, navigation elements, and generic footer content to ensure only substantive information is retained.

\section{Prompts Used}
\label{sec:appendix-prompts}

This section presents the complete prompts used by each agent in the \villainemoji~VILLAIN pipeline, together with concrete examples drawn from actual system outputs.

\begin{figure*}[ht]
\begin{tcolorbox}[colback=white, fontupper=\small]
\textbf{\# Role}\\
You are an expert AI Fact-Checker. Your specific task is to verify the \textbf{textual assertions} of a claim using the provided text-based sources.\\\\
\textbf{\# Input Data}\\
\textbf{\#\# 1. The Claim (Target for Verification)}\\
- \textbf{Claimant (Speaker):} \textcolor{gray}{Unknown}\\
- \textbf{Claim Date:} \textcolor{gray}{2024-01-15}\\
- \textbf{Claim Text:} \textcolor{gray}{Photo shows a 101-year-old woman who has given birth to her 17th child.}\\
- \textbf{Claim Images:} \textcolor{gray}{[Image of elderly woman holding baby]}\\\\
\textit{(Note: Use images for context, but focus verification on the text.)}\\\\
\textbf{\#\# 2. Retrieved Evidence}\\
- \textbf{Retrieved Text Sources:}\\
\textcolor{gray}{[1] (snopes.com) On 7 January 2016, the web site World News Daily Report published an article reporting that a 101-year-old Italian woman gave birth... As with all of World News Daily Report's content, the claim was completely fabricated. The article swiped an unrelated image from a Reddit thread...}\\
\textcolor{gray}{[2] (thequint.com) The 101-year-old woman seen in the photo is Rosa Camfield, holding her great-granddaughter Kaylee, and not her own child as claimed. Camfield passed away in 2015...}\\
\textcolor{gray}{[3] (vishvasnews.com) Vishvas News investigated and found that the claim is false... The article headlined `101-Year-Old Great-Grandmother in Heartwarming Viral Photo Dies' carries the same image...}\\\\
\textbf{\# Instructions}\\
1. \textbf{Contextual Understanding:} Analyze the Claim Text in conjunction with the Claim Images to fully understand the user's intent.\\
2. \textbf{Textual Verification:} Compare the factual claims made in the \textbf{Claim Text} against the \textbf{Retrieved Text Sources}. Look for: Factual support (dates, names, events); Contradictions or logical fallacies.\\
3. \textbf{Identify Information Gaps:} Explicitly state what information is missing from the \textit{sources} that is needed to verify the text claim fully.\\\\
\textbf{\# Output Format}\\
\textbf{\#\# 1. Key Verification Facts}\\
* [Fact]: (Evidence from sources supporting/refuting the text claim)\\\\
\textbf{\#\# 2. Missing Information}\\
* [Gap]: (Crucial info missing from sources)\\\\
\textbf{\#\# 3. Analysis}\\
(Summary of how text sources align with the claim text)\\\\
---\\
\textbf{Model Output:}\\
\textcolor{blue}{\textbf{\#\# 1. Key Verification Facts}}\\
\textcolor{blue}{* [Fact]: The woman in the photograph is Rosa Camfield, who was 101 years old when the photo was taken. The baby is her great-granddaughter, Kaylee. (The Quint, Vishvas News)}\\
\textcolor{blue}{* [Fact]: The story of a 101-year-old Italian woman named Anatolia Vertadella giving birth originated from a website called ``World News Daily Report,'' which has a disclaimer stating its articles are satirical and their content is fictional. (Snopes)}\\
\textcolor{blue}{* [Fact]: The official Guinness World Record for the oldest mother is Maria del Carmen Bousada Lara, who gave birth at age 66, making the claim of a 101-year-old giving birth medically implausible. (World Record Academy, CNN)}\\
\textcolor{blue}{* [Fact]: The woman in the photo, Rosa Camfield, passed away in March 2015, shortly after the photo went viral. (The Quint, Vishvas News)}\\\\
\textcolor{blue}{\textbf{\#\# 2. Missing Information}}\\
\textcolor{blue}{* The provided sources are sufficient to verify the claim. They identify the origin of the false story and the true context of the photograph.}\\\\
\textcolor{blue}{\textbf{\#\# 3. Analysis}}\\
\textcolor{blue}{The claim that the photo shows a 101-year-old woman who has given birth to her 17th child is \textbf{false}. The image is authentic but miscaptioned---it shows 101-year-old Rosa Camfield holding her two-week-old great-granddaughter, Kaylee, in 2015. The narrative of ``Anatolia Vertadella'' is a piece of fiction from the satirical website ``World News Daily Report.''}
\end{tcolorbox}
\caption{Prompt used for $\mathcal{A}_{TT}$ with an example of the input and output. Blue text indicates actual model output.}
\label{fig:agent1-prompt}
\end{figure*}

\begin{figure*}[ht]
\begin{tcolorbox}[colback=white, fontupper=\small]
\textbf{\# Role}\\
You are an expert AI Fact-Checker. Your specific task is to verify the \textbf{visual content} of the claim using the provided text-based sources.\\\\
\textbf{\# Input Data}\\
\textbf{\#\# 1. The Claim (Target for Verification)}\\
- \textbf{Claimant (Speaker):} \textcolor{gray}{Unknown}\\
- \textbf{Claim Date:} \textcolor{gray}{2024-01-15}\\
- \textbf{Claim Text:} \textcolor{gray}{Photo shows a 101-year-old woman who has given birth to her 17th child.}\\
- \textbf{Claim Images:} \textcolor{gray}{[Image of elderly woman holding baby]}\\\\
\textit{(Note: Use claim text to understand what the image purports to show.)}\\\\
\textbf{\#\# 2. Retrieved Evidence}\\
- \textbf{Retrieved Text Sources:}\\
\textcolor{gray}{[1] (africacheck.org) The photo shows Rosa Camfield, 101, cradling her newborn great-granddaughter. Camfield died just weeks later, on 30 March...}\\
\textcolor{gray}{[2] (chequeado.com) La foto corresponde a una mujer con su bisnieta y no es actual sino de 2015. La verdadera mujer de la foto es Rosa Camfield...}\\
\textcolor{gray}{[3] (snopes.com) World News Daily Report regularly publishes outlandish and false claims alongside repurposed photographs...}\\\\
\textbf{\# Instructions}\\
1. \textbf{Visual Analysis:} Analyze the visual elements in the \textbf{Claim Images} (landmarks, people, signs, weather).\\
2. \textbf{Cross-Modal Verification:} Check if the events or descriptions in the \textbf{Retrieved Text Sources} explain or contradict the visual elements.\\
\hspace{1em}\textit{Example:} Does the text report mention the specific objects or environment seen in the images?\\
3. \textbf{Identify Gaps:} What visual details are not explained by the text sources?\\\\
\textbf{\# Output Format}\\
\textbf{\#\# 1. Visual-Text Corroboration}\\
* [Point]: (How text sources confirm/deny specific visual elements)\\\\
\textbf{\#\# 2. Missing Context}\\
* [Gap]: (Visual details not mentioned in the text sources)\\\\
\textbf{\#\# 3. Analysis}\\
(Summary of the consistency between the image and the text reports)\\\\
---\\
\textbf{Model Output:}\\
\textcolor{blue}{\textbf{\#\# 1. Visual-Text Corroboration}}\\
\textcolor{blue}{* [Confirmed Visual Element - Elderly Woman]: The image shows a very elderly woman, consistent with the claim's description of a 101-year-old. Sources (The Quint, Snopes, Africa Check, Chequeado) confirm the woman is Rosa Camfield, who was 101 years old when the photo was taken in 2015.}\\
\textcolor{blue}{* [Confirmed Visual Element - Newborn Baby]: The image clearly shows a newborn baby. Sources confirm the baby is Kaylee, the woman's great-granddaughter, who was two weeks old at the time.}\\
\textcolor{blue}{* [Contradicted Relationship]: The claim states the woman is the baby's mother. Sources from Snopes, The Quint, Africa Check, and Chequeado all state this is false---Rosa Camfield is the baby's great-grandmother.}\\
\textcolor{blue}{* [Contradicted Narrative]: The story about ``Anatolia Vertadella'' giving birth is identified by Snopes and Chequeado as a complete fabrication from ``World News Daily Report,'' a known satirical/fake news website.}\\\\
\textcolor{blue}{\textbf{\#\# 2. Missing Context}}\\
\textcolor{blue}{* [True Identity]: The claim omits that the woman is Rosa Camfield and the baby is her great-granddaughter, Kaylee.}\\
\textcolor{blue}{* [Original Context]: The photo was taken in March 2015 in the United States and was shared by the baby's mother, Sarah Hamm, capturing the moment the 101-year-old met her great-granddaughter for the first time.}\\\\
\textcolor{blue}{\textbf{\#\# 3. Analysis}}\\
\textcolor{blue}{The visual elements superficially match the claim's description, but the sources reveal the image has been entirely decontextualized. The photograph shows a touching moment between a great-grandmother and her great-granddaughter, not a post-birth scene. The visual content does not support the narrative presented in the claim.}
\end{tcolorbox}
\caption{Prompt used for $\mathcal{A}_{IT}$ with an example of the input and output. Blue text indicates actual model output.}
\label{fig:agent2-prompt}
\end{figure*}

\begin{figure*}[ht]
\begin{tcolorbox}[colback=white, fontupper=\small]
\textbf{\# Role}\\
You are an expert AI Fact-Checker. Your specific task is to analyze the \textbf{cross-modal relationships} of the claim using the provided text and image sources.\\\\
\textbf{\# Input Data}\\
\textbf{\#\# 1. The Claim (Target for Verification)}\\
- \textbf{Claimant (Speaker):} \textcolor{gray}{Unknown}\\
- \textbf{Claim Date:} \textcolor{gray}{2024-01-15}\\
- \textbf{Claim Text:} \textcolor{gray}{Photo shows a 101-year-old woman who has given birth to her 17th child.}\\
- \textbf{Claim Images:} \textcolor{gray}{[CLAIM\_IMG\_1]}\\\\
\textbf{\#\# 2. All Retrieved Evidence}\\
- \textbf{Retrieved Text Sources:}\\
\textcolor{gray}{[1] (snopes.com) On 7 January 2016, the web site World News Daily Report published an article...}\\
\textcolor{gray}{[2] (thequint.com) The 101-year-old woman seen in the photo is Rosa Camfield...}\\
\textcolor{gray}{[3] (worldrecordacademy.org) Anatolia Vertadella, 101 year old, has become the oldest mother in history...}\\\\
- \textbf{Retrieved Source Images:} \textcolor{gray}{[RETRIEVED\_IMG\_1], [RETRIEVED\_IMG\_2], ...}\\\\
\textit{(Note: [CLAIM\_IMG\_n] tags represent claim images, and [RETRIEVED\_IMG\_n] tags represent retrieved source images. Treat these tags as placeholders for the actual visual data.)}\\\\
\textbf{\# Instructions}\\
1. \textbf{Source-to-Source Text Analysis:} Compare the retrieved text sources. Do they agree on key facts (dates, locations, names)? Identify any contradictions between sources.\\
2. \textbf{Cross-Modal Alignment:} Analyze if the \textbf{Retrieved Source Images} align with the narratives in the \textbf{Retrieved Text Sources}.\\
\hspace{1em}\textit{Example:} If Text Source A describes a ``sunny protest,'' does Image Source B show a sunny environment?\\
3. \textbf{Global Narrative Reconstruction:} Synthesize a coherent timeline or event description based on \textit{all} available evidence.\\
4. \textbf{Reliability Assessment:} Identify if any source seems like an outlier or low-quality compared to others.\\\\
\textbf{\# Output Format}\\
\textbf{\#\# 1. Evidence Consistency Check}\\
* [Text-Text]: (Do text sources agree? Note contradictions.)\\
* [Image-Text]: (Do source images support the source texts?)\\
* [Image-Image]: (Is there visual consistency between the claim image and sources, and among sources themselves?)\\\\
\textbf{\#\# 2. Global Context Summary}\\
(A unified summary of the event based on the combined evidence, independent of the user's claim)\\\\
\textbf{\#\# 3. Conflict Alert}\\
* [Conflict]: (Critical discrepancies between sources, if any)\\\\
---\\
\textbf{Model Output:}\\
\textcolor{blue}{\textbf{\#\# 1. Evidence Consistency Check}}\\
\textcolor{blue}{* [Text-Text]: There is a major contradiction. Group 1 (World Record Academy, guardian.ng, etc.) repeats the story of ``Anatolia Vertadella'' giving birth. Group 2 (Snopes, The Quint, Vishvas News, Africa Check) debunks this as a fabrication from ``World News Daily Report,'' a satirical website. The fact-checkers identify the woman as Rosa Camfield and the baby as her great-granddaughter.}\\
\textcolor{blue}{* [Image-Text]: Direct conflict. Group 1 mislabels the image as ``Anatolia Vertadella'' with her newborn son. Group 2 correctly identifies it as Rosa Camfield meeting her great-granddaughter.}\\
\textcolor{blue}{* [Image-Image]: The claim image is visually identical to retrieved images, confirming all sources discuss the same photograph.}\\\\
\textcolor{blue}{\textbf{\#\# 2. Global Context Summary}}\\
\textcolor{blue}{In March 2015, a photograph was taken of 101-year-old Rosa Camfield holding her two-week-old great-granddaughter, Kaylee. The photo went viral. In January 2016, the satirical website ``World News Daily Report'' created a fabricated story about ``Anatolia Vertadella'' giving birth at 101, using this unrelated photo. Multiple fact-checking organizations have debunked this story.}\\\\
\textcolor{blue}{\textbf{\#\# 3. Conflict Alert}}\\
\textcolor{blue}{* [Conflict]: Critical conflict between the claim and verified facts. Identity Misrepresentation: The woman is Rosa Camfield (American), not ``Anatolia Vertadella'' (Italian). Relationship Misrepresentation: The baby is her great-granddaughter, not her child. Event Fabrication: The birth story is fiction from a satirical news source.}
\end{tcolorbox}
\caption{Prompt used for $\mathcal{A}_{CM}$ with an example of the input and output. Blue text indicates actual model output.}
\label{fig:agent3-prompt}
\end{figure*}

\begin{figure*}[ht]
\begin{tcolorbox}[colback=white, fontupper=\small]
\textbf{\# Role}\\
You are the Lead Fact-Checking Adjudicator. Your task is to synthesize preliminary analyses into \textbf{decisive Question-Answer (QA) pairs} that consolidate the key evidence and reasoning required to form a final verdict.\\\\
\textbf{\# Input Data}\\
\textbf{\#\# 1. The Claim (Target for Verification)}\\
- \textbf{Claimant (Speaker):} \textcolor{gray}{Unknown}\\
- \textbf{Claim Date:} \textcolor{gray}{2024-01-15}\\
- \textbf{Claim Text:} \textcolor{gray}{Photo shows a 101-year-old woman who has given birth to her 17th child.}\\
- \textbf{Claim Images:} \textcolor{gray}{[Image attached]}\\\\
\textbf{\#\# 2. Preliminary Analyses}\\
\textcolor{gray}{[Agent 1 Output]: Key Verification Facts: The woman in the photograph is Rosa Camfield... The story originated from World News Daily Report...}\\
\textcolor{gray}{[Agent 2 Output]: Visual-Text Corroboration: The image shows a very elderly woman with a newborn baby... Missing Context: True identity of the individuals...}\\
\textcolor{gray}{[Agent 3 Output]: Evidence Consistency Check: Text-Text shows major contradiction... Global Context Summary: In March 2015, Rosa Camfield...}\\\\
\textbf{\#\# 3. Few-shot Learning Examples}\\
\textcolor{gray}{(Retrieved via BM25 from training set based on claim similarity)}\\\\
\textbf{\#\# 4. Previously Generated QA Pairs}\\
\textit{(Do NOT repeat these or ask similar questions)}\\
\textcolor{gray}{[Previously generated pairs listed here in subsequent iterations]}\\\\
\textbf{\# Instructions}\\
\textbf{\#\# Synthesize Diagnostic QAs (The Reasoning Basis)}\\
Analyze the provided forensic reports to extract the \textbf{core information} necessary to predict the verdict. Formulate \textbf{N high-impact QA pairs} that:\\
- \textbf{Isolate Key Evidence:} Focus on dates, locations, inconsistencies, or manipulation traces that act as ``smoking guns.''\\
- \textbf{Resolve Ambiguity:} Ask and answer questions that clarify whether the evidence is sufficient or conflicting.\\
- \textbf{Serve as Proof:} Each QA must act as a logical premise supporting your final decision.\\\\
\textbf{\# Output Format (JSON Only)}\\
\texttt{\{"qa\_pairs": [\{"question": "<Question 1>", "answer": "<Full statement answer 1>"\}, \{"question": "<Question 2>", "answer": "<Full statement answer 2>"\}]\}}\\\\
---\\
\textbf{Model Output:}\\
\textcolor{blue}{\{"qa\_pairs": [}\\
\textcolor{blue}{~~\{"question": "Who are the people in the photograph and what is their actual relationship?",}\\
\textcolor{blue}{~~~"answer": "The photograph shows 101-year-old Rosa Camfield holding her great-granddaughter, Kaylee, in March 2015. The baby is not her child."\},}\\
\textcolor{blue}{~~\{"question": "Is it medically plausible for a 101-year-old to give birth?",}\\
\textcolor{blue}{~~~"answer": "No, giving birth at 101 is considered medically implausible. The verified Guinness World Record for the oldest mother is Maria del Carmen Bousada Lara, who gave birth at the age of 66."\},}\\
\textcolor{blue}{~~\{"question": "What is the origin of the story about a 101-year-old Italian woman named Anatolia Vertadella giving birth?",}\\
\textcolor{blue}{~~~"answer": "The story is a complete fabrication that originated from 'World News Daily Report,' a website known for publishing satirical and fictional content."\},}\\
\textcolor{blue}{~~\{"question": "When did the woman in the photo, Rosa Camfield, pass away?",}\\
\textcolor{blue}{~~~"answer": "Rosa Camfield passed away in March 2015, shortly after the photo was taken, which was years before many versions of the false claim began circulating."\}]\}}
\end{tcolorbox}
\caption{Prompt used for $\mathcal{A}_{QA}$ with an example of the input and output. Blue text indicates actual model output in JSON format.}
\label{fig:agent4-prompt}
\end{figure*}

\begin{figure*}[ht]
\begin{tcolorbox}[colback=white, fontupper=\small]
\textbf{\# Role}\\
You are the Lead Fact-Checking Adjudicator. Your task is to select the most relevant QA pairs, assess veracity, and provide a final verdict with justification.\\\\
\textbf{\# Input Data}\\
\textbf{\#\# 1. The Claim (Target for Verification)}\\
- \textbf{Claimant (Speaker):} \textcolor{gray}{Unknown}\\
- \textbf{Claim Date:} \textcolor{gray}{2024-01-15}\\
- \textbf{Claim Text:} \textcolor{gray}{Photo shows a 101-year-old woman who has given birth to her 17th child.}\\
- \textbf{Claim Images:} \textcolor{gray}{[Image attached]}\\\\
\textbf{\#\# 2. Generated Question-Answer Pairs}\\
\textcolor{gray}{1. \textbf{Q:} Who are the people in the photograph?}\\
\textcolor{gray}{~~~\textbf{A:} The photograph shows 101-year-old Rosa Camfield holding her great-granddaughter, Kaylee, in March 2015.}\\
\textcolor{gray}{2. \textbf{Q:} What is the origin of the story about Anatolia Vertadella?}\\
\textcolor{gray}{~~~\textbf{A:} The story is a complete fabrication from ``World News Daily Report,'' a satirical website.}\\
\textcolor{gray}{3. \textbf{Q:} Is it medically plausible for a 101-year-old to give birth?}\\
\textcolor{gray}{~~~\textbf{A:} No. The Guinness World Record for oldest mother is 66 years old.}\\
\textcolor{gray}{4. \textbf{Q:} What happened to Rosa Camfield after the photo was taken?}\\
\textcolor{gray}{~~~\textbf{A:} Rosa Camfield passed away on March 30, 2015, just weeks after meeting her great-granddaughter.}\\
\textcolor{gray}{...}\\\\
---\\
\textbf{\# Instructions}\\
1. \textbf{Select Best QA Pairs:} From the generated QA pairs above, select the \textbf{N most relevant and informative} pairs for verification.\\
2. \textbf{Determine Verdict:} Choose the single best label:\\
\hspace{1em}- \textbf{Supported}\\
\hspace{1em}- \textbf{Refuted}\\
\hspace{1em}- \textbf{Not Enough Evidence}\\
\hspace{1em}- \textbf{Conflicting Evidence/Cherrypicking}\\
3. \textbf{Write Justification:} A cohesive summary explaining the verdict based on the selected QA pairs.\\
4. \textbf{JSON Output:} Output \textbf{ONLY} a valid JSON object matching the format below.\\\\
\textbf{\# Output Format (JSON Only)}\\
\texttt{\{"questions": [\{"question": "<Selected question 1>", "answer": "<Answer 1>"\}, ...], "veracity\_verdict": "<Label>", "justification": "<Explanation>"\}}\\\\
---\\
\textbf{Model Output:}\\
\textcolor{blue}{\{}\\
\textcolor{blue}{~~"questions": [}\\
\textcolor{blue}{~~~~\{"question": "Who are the people in the photograph and what is their actual relationship?",}\\
\textcolor{blue}{~~~~~"answer": "The photograph shows 101-year-old Rosa Camfield holding her great-granddaughter, Kaylee, in March 2015. The baby is not her child."\},}\\
\textcolor{blue}{~~~~\{"question": "What is the origin of the story about a 101-year-old Italian woman named Anatolia Vertadella giving birth?",}\\
\textcolor{blue}{~~~~~"answer": "The story is a complete fabrication that originated from 'World News Daily Report,' a website known for publishing satirical and fictional content."\},}\\
\textcolor{blue}{~~~~\{"question": "Is it medically plausible for a 101-year-old to give birth?",}\\
\textcolor{blue}{~~~~~"answer": "No, giving birth at 101 is considered medically implausible. The verified Guinness World Record for the oldest mother is Maria del Carmen Bousada Lara, who gave birth at the age of 66."\}],}\\
\textcolor{blue}{~~"veracity\_verdict": "Refuted",}\\
\textcolor{blue}{~~"justification": "The claim is a complete fabrication. The photograph does not show a mother and her newborn child, but rather 101-year-old Rosa Camfield holding her great-granddaughter, Kaylee, in March 2015. This was confirmed by her family, who originally shared the photo. Furthermore, the woman in the photo, Rosa Camfield, passed away in 2015, making it impossible for her to be the subject of a story that began circulating widely in 2016. The narrative of a 101-year-old Italian woman giving birth originated from 'World News Daily Report,' a well-known satirical website that explicitly states its content is fictional. The claim is also medically implausible, as the confirmed world record for the oldest mother is 66, not 101."}\\
\textcolor{blue}{\}}
\end{tcolorbox}
\caption{Prompt used for $\mathcal{A}_{V}$ with an example of the input and output. Blue text indicates the actual model output, including the final verdict of Refuted.}
\label{fig:agent5-prompt}
\end{figure*}

\end{document}